\documentclass{article}
\usepackage{spconf,amsmath,graphicx,hyperref}
\usepackage{array}
\usepackage{algorithm}
\usepackage{algorithmic}
\usepackage{multirow}
\usepackage{subfigure}
\usepackage{xcolor}
\usepackage{colortbl}

\title{Federated Heterogeneous Language Model Optimization for Hybrid Automatic Speech Recognition}

\name{
\vspace{-2px}
\begin{tabular}{c}
Mengze Hong$  ^{1*}  $ \qquad
Yi Gu$  ^{2,3*}  $ \qquad
Di Jiang$  ^{1\dagger}  $ \qquad
Hanlin Gu$  ^{4}  $ \\ \vspace{0px}
\textit{Chen Jason Zhang}$  ^{1}  $ \qquad
\textit{Lu Wang}$  ^{5}  $ \qquad
\textit{Zhiyang Su$  ^{4}  $}
\end{tabular}
\thanks{$^{*}$Equal Contribution. $^{\dagger}$Corresponding Author.}}

\address{
$^{1}$Hong Kong Polytechnic University \quad
$^{2}$Ant Group \quad
$^{3}$Shanghai Jiao Tong University \\
$^{4}$Hong Kong University of Science and Technology \quad
$^{5}$Shenzhen University 
}

\begin{document}
\ninept
\maketitle
\begin{abstract}
Training automatic speech recognition (ASR) models increasingly relies on decentralized federated learning to ensure data privacy and accessibility, producing multiple local models that require effective merging. In hybrid ASR systems, while acoustic models can be merged using established methods, the language model (LM) for rescoring the N-best speech recognition list faces challenges due to the heterogeneity of non-neural n-gram models and neural network models. This paper proposes a heterogeneous LM optimization task and introduces a match-and-merge paradigm with two algorithms: the Genetic Match-and-Merge Algorithm (GMMA), using genetic operations to evolve and pair LMs, and the Reinforced Match-and-Merge Algorithm (RMMA), leveraging reinforcement learning for efficient convergence. Experiments on seven OpenSLR datasets show RMMA achieves the lowest average Character Error Rate and better generalization than baselines, converging up to seven times faster than GMMA, highlighting the paradigm's potential for scalable, privacy-preserving ASR systems.
\end{abstract}

\section{Introduction}

Automatic speech recognition (ASR) is a core technology that enables accessible and scalable speech-driven applications \cite{moriya2025alignment, lu2025contextualizedtokendiscriminationspeech}. Training high-performance ASR systems requires large volumes of carefully curated speech data, which are increasingly managed through decentralized federated learning due to growing concerns about data privacy \cite{10.1145/3298981}. In this paradigm, data curators train local models on private datasets and share only model parameters to construct a stronger global model, enabling collaborative optimization without direct data exposure \cite{gao2022end}. Despite its effectiveness, federated ASR training requires tailored optimization of the model aggregation process to achieve performance comparable to centralized training without incurring excessively high computational cost \cite{10.1007/978-3-030-59419-0_54}.

In practice, hybrid ASR systems are widely used due to their modular design and support for real-time streaming \cite{le9054257,zei9746377}, integrating acoustic models (AMs) and language models (LMs) to achieve robust performance. AMs extract features from raw speech signals and model the relationship between acoustic feature vectors and phonetic units, converting audio into phone sequences \cite{dong2018speech}. LMs estimate word sequence probabilities and rescore the N-best hypothesis list to incorporate long-range semantic information. A typical LM pipeline employs an $n$-gram model to generate an N-best list or lattice, followed by a neural network (NN) model that rescores candidate sentences to produce the final recognition output \cite{2018Rescoring,liu9414806}. Existing research has primarily focused on acoustic model optimization, with methods such as GMA and SOMA enabling effective aggregation \cite{ijcai2020-513, tan2020federated}. In contrast, language model optimization remains underexplored and faces two significant technical barriers:

\begin{enumerate}
\item \textbf{Heterogeneity}: The n-gram model differs fundamentally in structure from neural network models, making existing optimization methods designed for isomorphic models infeasible.
\item \textbf{Alignment}: Effective LM merging requires mutual matching over the N-best list, as independent optimization of individual LMs does not guarantee optimal combined performance.
\end{enumerate}

\begin{figure}
    \centering
	\includegraphics[width=\linewidth]{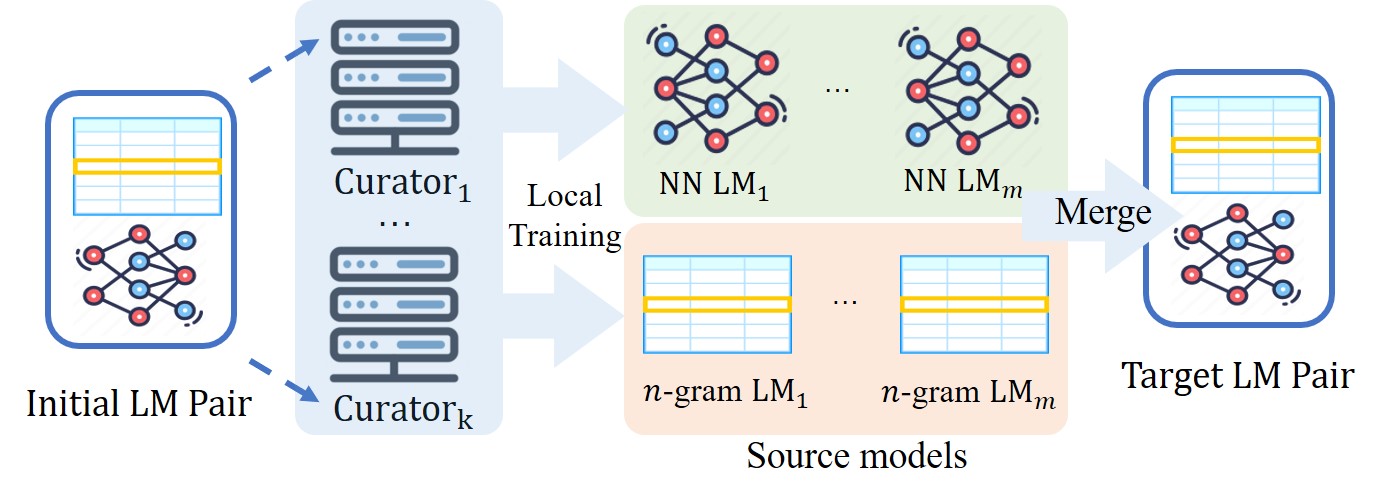}
    \vspace{-2.2em}
	\caption{The optimization of heterogeneous language model (LM) pairs in ASR involves (1) local training of source model pairs by curators using private data, and (2) merging multiple models to form a superior target model pair. The values of $m$ and $k$ may differ due to differences in data distribution across curators.}
    \vspace{-1em}
	\label{fig:teaser}
\end{figure}

\noindent To overcome these challenges, this paper introduces a novel heterogeneous language model optimization task for federated hybrid ASR systems. We present a unified \textbf{match-and-merge paradigm}, using genetic algorithms to handle n-gram and neural network LMs as distinct populations. Based on this paradigm, we develop two innovative algorithms and demonstrate their superior performance and convergence efficiency compared to existing methods.

\begin{figure*}
\centering
\includegraphics[width=0.67\linewidth]{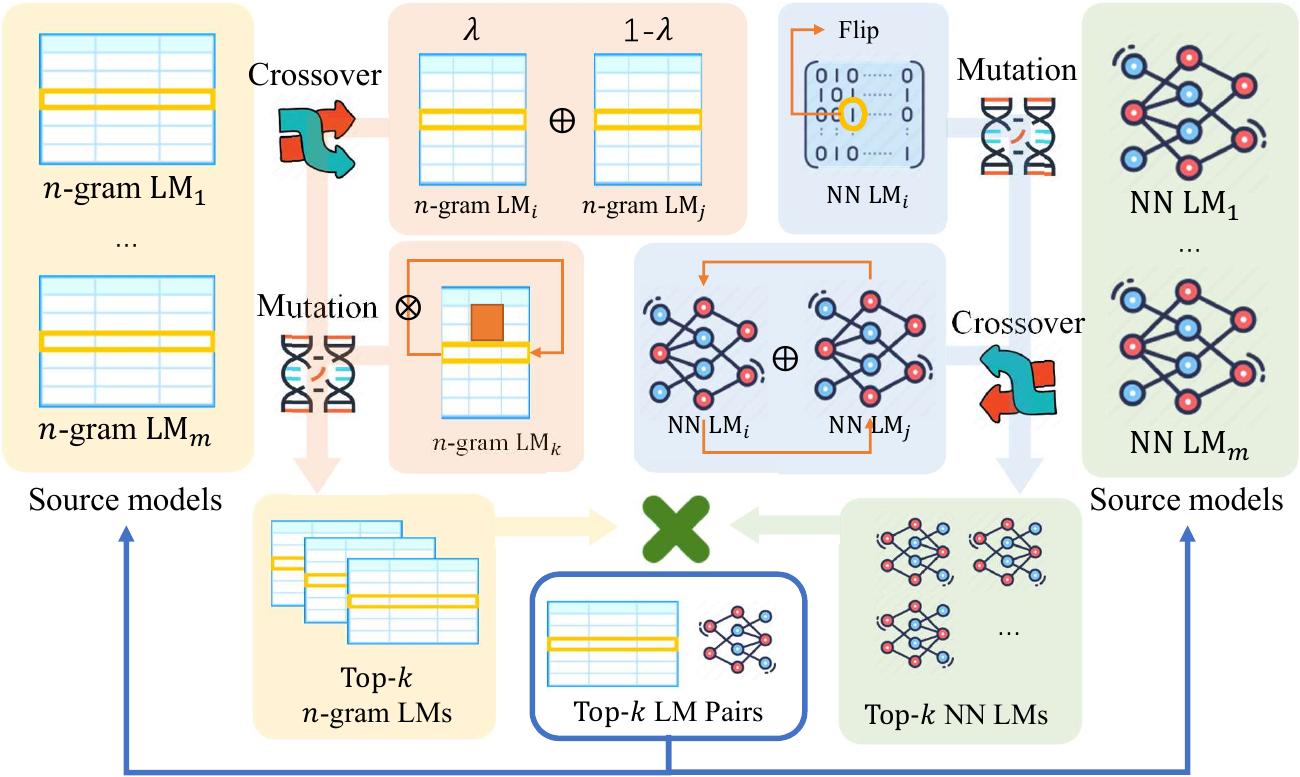}
\vspace{-0.5em}
\caption{Overview of the Genetic Match-and-Merge Algorithm (GMMA). Heterogeneous language models (n-gram and neural LMs) are treated as separate populations and evolved via type-specific genetic operations. The top-$k$ candidates from each population are then paired, and LM combinations with the highest matching fitness are selected for reproduction and merging.}
\label{gmma}
\end{figure*}

\section{Related Work}

Hybrid ASR systems, which combine acoustic models and language models, have evolved rapidly from traditional GMM-HMM architectures \cite{swietojanski2013revisiting} to deep learning-based frameworks \cite{kheddar2024automatic}. This hybrid design offers several advantages over end-to-end systems: superior explainability for error analysis, modularity that allows AMs and LMs to be independently tuned or adapted to specific domains and datasets, and natural support for decentralized, heterogeneous data, enabling personalized model adaptation \cite{rouhe2022low}. These features have made hybrid ASR systems widely adopted in industrial applications through toolkits such as Kaldi ASR, which facilitate lightweight and efficient model training and deployment \cite{perero2022comparison}.

Although heterogeneous language model optimization has received limited attention, many studies focus on model combination (or ensembling), where each model predicts independently, and outputs are later aggregated \cite{zhou2012ensemble, parikh2024ensembles}. In contrast, model merging targets training and model-level optimization, integrating parameters from locally trained models into a single strong model of the same size \cite{ijcai2020-513}. This approach is better suited to large-scale industrial applications, producing a single, optimized model that reduces inference cost and latency while maintaining robustness across datasets.

\section{Problem Formulation}

Consider $n$ private datasets $T = \{T_1, \dots, T_n\}$, each corresponding to a distinct application scenario or data distribution, and used to train a paired source model $M_i=(M_{N,i}, M_{R,i})$, where $M_{N,i}$ is the $i$-th $n$-gram LM and $M_{R,i}$ is the $i$-th neural network language model (NN LM). The $n$-gram LMs share a homogeneous structure with distinct parameters, as do the NN LMs, but $n$-gram and NN LMs are heterogeneous. Each $n$-gram LM is represented as an $n$-dimensional matrix of word sequence frequencies, with column vectors $A_{N,i}^j$ ($1 \le j \le n$). Each NN LM $M_{R,i}$ consists of $L$ layers, with parameters $W_{R,i}^l$ ($1 \le l \le L$) representing weights and biases. The optimization task is to combine the $n$ paired source models into a target model pair $M_T=(M_{N,T}, M_{R,T})$ through weighted parameter combination of $M_{N,i}$ and $M_{R,i}$.

\begin{algorithm}[!t]
\caption{Genetic Match-and-Merge Algorithm (GMMA)}
\label{algorithm1}
\textbf{Input}: Source paired models $M_1, M_2, \dots, M_n$ \\
\textbf{Initialize} $P = \{M_1,\dots, M_n\}$, $P_N = \{M_{N,1}, \dots, M_{N,n}\}$, $P_R = \{M_{R,1}, \dots, M_{R,n}\}$
\begin{algorithmic}[1]
\WHILE {not converged}
\FOR {each $M_{N,i}$ in $P_N$}
\STATE {With $p_1$, $P_N \gets P_N \cup \text{Mutation}(M_{N,i})$}
\ENDFOR
\FOR {each $M_{R,i}$ in $P_R$}
\STATE {With $p_1$, $P_R \gets P_R \cup \text{Mutation}(M_{R,i})$}
\ENDFOR
\STATE Randomly shuffle $P$
\FOR {each adjacent pair $(M_{N,i}, M_{N,i+1})$ in $P_N$}
\STATE {With $p_2$, $P_N \gets P_N \cup \text{Crossover}(M_{N,i}, M_{N,i+1})$}
\ENDFOR
\FOR {each adjacent pair $(M_{R,i}, M_{R,i+1})$ in $P_R$}
\STATE {With $p_2$, $P_R \gets P_R \cup \text{Crossover}(M_{R,i}, M_{R,i+1})$}
\ENDFOR
\STATE Reproduction via pairing $P_N \times P_R$
\STATE Set $P$ as the top-$K$ pairs
\ENDWHILE
\end{algorithmic}
\end{algorithm}

\section{Genetic Match-and-Merge Algorithm}
We propose the \textbf{Genetic Match-and-Merge Algorithm} (GMMA) to optimize the merging of heterogeneous $n$-gram and NN LMs (see Figure \ref{gmma}). GMMA leverages a genetic algorithm inspired by natural selection \cite{sampson1976adaptation}, starting from the source models and evolving over generations through reproduction, mutation, and crossover operators to produce increasingly well-adapted offspring. For clarity, let $P$ denote the overall population of paired models, while $P_N$ and $P_R$ include only the $n$-gram and NN LMs, respectively.

The initial population consists of the source models, with genetic operators specifically designed for heterogeneous LMs. For NN LMs, mutation randomly flips a bit in the model’s binary file, while crossover selects two adjacent models and exchanges layers at a random point $l$ ($1 \le l \le L$) to generate two offspring \cite{mcmahan2017communication}:

\vspace{-1em}
{\ninept
\begin{equation*}
\begin{aligned}
&M_{R,i'} = \{W_{R,i'}^1, \dots, W_{R,i'}^l, W_{R,i+1'}^{l+1}, \dots, W_{R,i+1'}^L\}, \\
&M_{R,i+1'} = \{W_{R,i+1'}^1, \dots, W_{R,i+1'}^l, W_{R,i'}^{l+1}, \dots, W_{R,i'}^L\}.
\end{aligned}
\label{eq:crossoverR}
\end{equation*}
}
\vspace{-0.5em}

\noindent For $n$-gram LMs, mutation scales a randomly selected column vector by a coefficient $k \in (0,1)$:

\vspace{-1em}
{\ninept
\begin{equation*}
M_{N,i'} = [0^1, \dots, 0^{j-1}, k \cdot A_{N,i}^j, 0^{j+1}, \dots, 0^n] + M_{N,i}.
\label{eq:mutation}
\end{equation*}
}
\vspace{-1em}

The crossover operation combines two adjacent models:

\vspace{-0.8em}
{\ninept
\begin{equation*}
M_{N,i'} = \lambda M_{N,i} + (1 - \lambda) M_{N,i+1},
\label{eq:crossoverN}
\end{equation*}
}
\vspace{-1.3em}

\noindent with $\lambda \in (0,1)$ representing a random weight. For reproduction, GMMA considers heterogeneous $n$-gram and NN LMs as separate populations, with fitness defined by their degree of match, evaluated using the Character Error Rate (CER) on a validation dataset. The top $K$ $n$-gram LMs are paired with the top $K$ NN LMs, and the $K$ pairs with the lowest CERs are selected as parents for the next generation. Increasing $K$ improves population diversity and expands the search space, but also increases computational costs. Algorithm \ref{algorithm1} presents GMMA, with hyperparameters $p_1$ and $p_2$ controlling the mutation and crossover probabilities, respectively.

\begin{algorithm}[!t]
\caption{Reinforced Match-and-Merge Algorithm (RMMA)}
\label{algorithm_rmma}
\textbf{Input}: Source paired models $M_1, M_2, \dots, M_n$

\textbf{Initialize}: policy network $\omega_a$, discount factor $\gamma$, learning rate $\beta$ 

\begin{algorithmic}[1]
\STATE Set $t \gets 1$, initialize target model state $s_t = (s_t^M, s_t^R)$
\REPEAT
    \IF{$\text{train}$}
        \STATE Sample action $a_t \sim \pi(a_t|s_t; \omega_a)$
    \ELSE
        \STATE Select action $a_t = \arg\max_a \pi(a|s_t; \omega_a)$
    \ENDIF
    \STATE Execute $a_t$, update merged model: $s_{t+1}^M = f(a_t; M_R, M_N)$
    \STATE Update observation $s_{t+1}^R$, state $s_{t+1} = (s_{t+1}^M, s_{t+1}^R)$
    \IF{$\text{train}$}
        \STATE Compute TD error: $\delta_t \gets r_t + \gamma \cdot v(s_{t+1}) - v(s_t)$
        \STATE Update policy: $\omega_a \gets \omega_a + \beta \cdot \nabla_{\omega_a} \log \pi(a_t|s_t; \omega_a) \cdot \delta_t$
    \ENDIF
    \STATE $t \gets t + 1$
\UNTIL{$t > t_\text{max}$ or $a_{t-1} = \text{terminate}$}
\STATE \textbf{Return} optimized model pair $M_t$
\end{algorithmic}
\end{algorithm}

\section{Reinforced Match-and-Merge Algorithm}

\begin{figure}[!t]
\centering
\includegraphics[width=\linewidth]{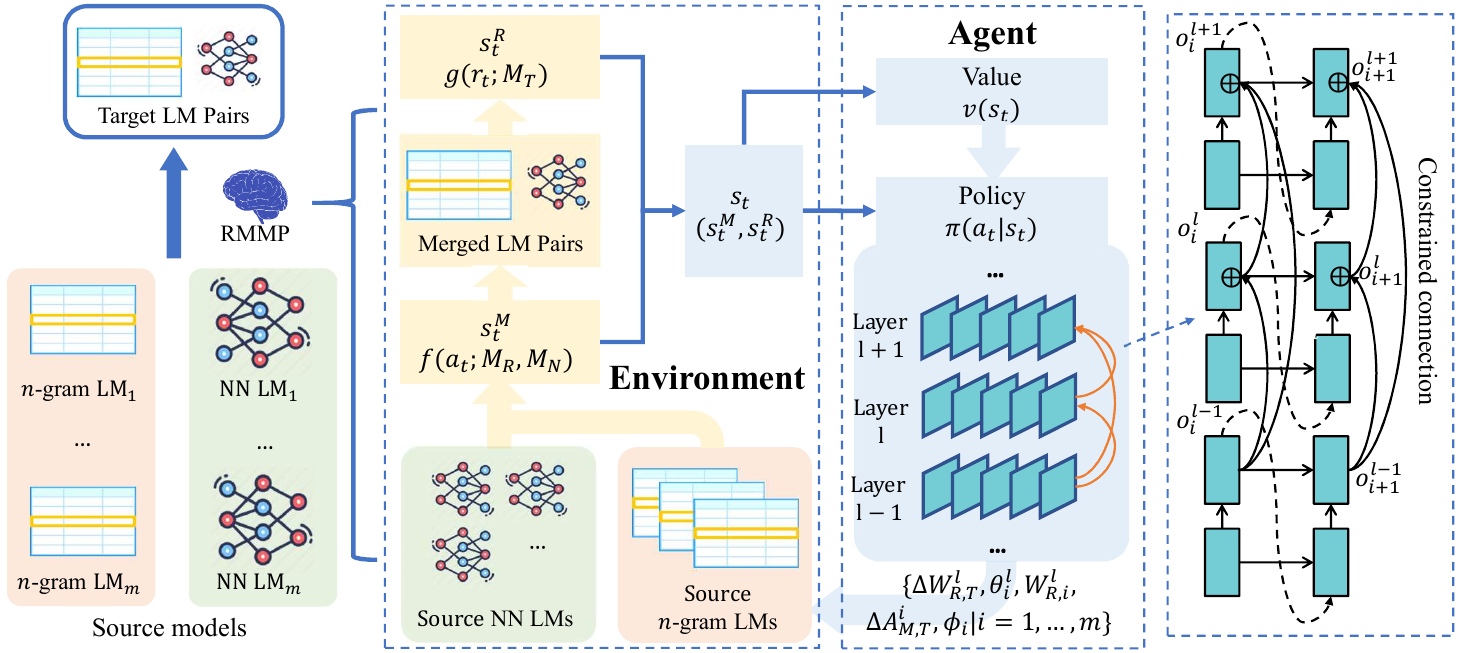}
\vspace{-1.5em}
\caption{Overview of Reinforced Match-and-Merge Algorithm. The merging problem is formulated as a sequential decision-making process, where a reinforcement learning agent selects actions to maximize rewards and enhance model quality.}
\vspace{-1.15em}
\label{rnn}
\end{figure}

Although the GMMA approach produces high-quality language models, its reliance on random crossover and mutation leads to slow convergence, limiting practical applicability. To achieve comparable performance more efficiently, we propose the \textbf{Reinforced Match-and-Merge Algorithm} (RMMA), which employs a reinforcement learning agent \cite{mnih2013playing} to guide the merging process (Algorithm~\ref{algorithm_rmma}). As illustrated in Figure~\ref{rnn}, the source models are treated as the environment, while the action set $\mathcal{A} = [a_1, \dots, a_t]$ specifies the merge variables. The state $s_t = (s_t^M, s_t^R)$ consists of the merged model pair $s_t^M = f(a_t; M_R, M_N)$ and the corresponding evaluation feedback $s_t^R$ obtained from a validation set. The merging of NN LMs is formulated by extending the rigorous proposition in \cite{ijcai2020-513}:
\begin{equation}
\setlength{\jot}{-0.2em}
\begin{aligned}
&W_{R,T}^{l} = \sum_{i=1}^{n} \theta_{i}^{l} W_{R,i}^{l} + \Delta W_{R,T}^{l} \
&\mathrm{s.t.} \quad \theta_{i}^{l} > 0,\ \sum_{i=1}^{n} \theta_{i}^{l} = 1,
\end{aligned}
\label{NN}
\end{equation}
\vspace{-0.6em}

\noindent where $\Delta W_{R,T}^{l}$ ($1 \leq l \leq L$) captures the perturbations introduced by the mutation operator, while the weighted summation term $\sum_{i=1}^{n} \theta_{i}^{l} W_{R,i}^{l}$ corresponds to crossover. For n-gram LMs, we define the merging analogously as:

\vspace{-0.6em}
\begin{equation}
\setlength{\jot}{-0.2em}
\begin{aligned}
&M_{N,T}=\sum_{i=1}^{n} \phi_{i} M_{N,i}+\sum_{j=1}^{n}\Delta A_{M,T}^{j}
&\mathrm{ s.t. }\quad \phi_{i}>0,\sum_{i=1}^{n} \phi_{i} =1,
\end{aligned}
\label{ngram}
\end{equation}
\vspace{-0.6em}

\noindent where $\Delta A_{M,T}^{j}$ ($1 \leq j \leq n$) denotes the mutation effects. With these definitions, the optimization objective is formulated as follows:

\begin{equation}
\begin{aligned}
\min_{\Delta W_{R,T}^{l},\,\theta_{i}^{l},\,\Delta A_{M,T}^{j},\,\phi_{i}}
\ell(M_{N,T}, M_{R,T}),
\end{aligned}
\label{optimization}
\end{equation}
\noindent subject to the constraints in Equations~\ref{NN} and~\ref{ngram}.

The reinforcement learning agent is implemented as an actor-critic model~\cite{konda2000actor}, using a recurrent neural network to parameterize the policy $\pi(a_t|s_t)$ for action selection. The policy is conditioned on the RNN hidden state $h_t$ at time step $t$:

\begin{equation}
    h_t = \tanh(W_p s_t + U_c h_{t-1}),
\end{equation}
where $W_p$ and $U_c$ are trainable parameters. Let $\ell(M)$ denote the CER on a validation set. The reward signal $r_t$ is computed as:
\begin{equation}
r_t = 
\begin{cases}
\eta \cdot \text{sign}[\ell(M_{t-1}) - \ell(M_{t})], & a_t \neq \text{terminate}, \\
\text{sign}[\epsilon - \ell(M_t)], & a_t = \text{terminate},
\end{cases}
\end{equation}
where $\eta$ is a scale factor and $\epsilon$ is a target CER threshold.

We train the agent using one-step temporal-difference (TD) learning~\cite{Sutton1988Learning}. With discount factor $\gamma \in [0,1)$, the TD error (used as an advantage estimate) is:
\begin{equation}
    \delta_t = r_t + \gamma \cdot v(s_{t+1}) - v(s_t).
\end{equation}
The policy gradient is approximated via the actor-critic update:
\begin{equation}
    \nabla_{\omega_a} J(\omega_a) \approx \nabla_{\omega_a} \log \pi(a_t|s_t; \omega_a) \cdot \delta_t,
\end{equation}
where $\omega_a$ are the policy parameters and $J(\omega_a)$ is the expected cumulative discounted reward. During inference, the agent selects actions greedily as $a_t = \arg\max_a \pi(a|s_t; \omega_a)$ and returns the final merged model pair upon termination.

\begin{table*}[!t]
\centering
\caption{Dataset statistics (sample size and duration) and evaluation results, reported as the mean over 5 random seeds. Lower CER indicates better performance; the best results among merged models are shown in \textbf{bold}.}
\vspace{-1em}
\resizebox{0.98\textwidth}{!}{
\begin{minipage}[t]{0.35\textwidth} 
\centering
\scriptsize
\renewcommand{\arraystretch}{1.2}
\vspace{1em}
\begin{tabular}{|c|c|c|c|}
\hline
\multirow{2}{*}{\textbf{Dataset}} & \multirow{2}{*}{\textbf{Name}} & \multicolumn{2}{c|}{\textbf{Data Statistics}} \\ \cline{3-4}
& & no.wav & hours \\ \hline
SLR18 & THCHS-30 & 13388 & 34.0 \\ \hline
SLR33 & Aishell & 141525 & 179.3 \\ \hline
SLR38 & Free ST Mandarin & 102600 & 109.5 \\ \hline
SLR47 & Primewords & 50384 & 98.8 \\ \hline
SLR62 & aidatatang & 237265 & 200.3 \\ \hline
SLR68 & Read Speech & 658084 & 768.8 \\ \hline
SLR93 & AISHELL-3 & 88035 & 85.1 \\ \hline
\textbf{Total} & -- & 1291281 & 1475.8 \\ \hline
\end{tabular}
\label{datasets}
\end{minipage}
\hfill
\begin{minipage}[t]{0.64\textwidth} 
\centering
\scriptsize
\renewcommand{\arraystretch}{1.2}
\vspace{1em}
\begin{tabular}{|l|c|c|c|c|c|>{\columncolor{gray!20}}c|>{\columncolor{gray!20}}c|>{\columncolor{gray!20}}c|}
\hline
\textbf{Model} & \textbf{SLR33} & \textbf{SLR38} & \textbf{SLR47} & \textbf{SLR62} & \textbf{SLR93} & \textbf{Average} & \textbf{SLR18} & \textbf{SLR68} \\ \hline
Reference & 4.60 & 8.91 & 17.08 & 4.20 & 4.63 & 7.88 & 13.47 & 8.53 \\ \hline
\multicolumn{9}{|c|}{\textbf{Source Models}} \\ \hline
SLR33 & 4.25 & 13.87 & 18.99 & 9.44 & 6.51 & 10.61 & 14.86 & 14.66 \\ \hline
SLR38 & 14.20 & 8.51 & 27.90 & 9.53 & 13.40 & 14.71 & 23.49 & 14.71 \\ \hline
SLR47 & 7.78 & 15.02 & 15.79 & 10.51 & 8.33 & 11.48 & 14.21 & 14.69 \\ \hline
SLR62 & 14.74 & 15.28 & 27.12 & 4.09 & 12.50 & 14.74 & 24.15 & 10.21 \\ \hline
SLR93 & 5.16 & 13.46 & 19.63 & 8.58 & 5.26 & 10.42 & 15.68 & 10.84 \\ \hline
\multicolumn{9}{|c|}{\textbf{Merged Models}} \\ \hline
Fine-tuning & 7.21 & 11.28 & 18.99 & 7.76 & 7.35 & 10.51 & 15.81 & 11.09 \\ \hline
Direct Average & 5.14 & 8.85 & 18.96 & 4.54 & 5.18 & 8.53 & 14.98 & 10.16 \\ \hline
GMMA & 4.63 & 11.33 & 17.57 & 4.17 & 4.57 & 8.46 & 14.08 & 8.81 \\ \hline
RMMA & 4.63 & 9.26 & 17.59 & 4.14 & 4.55 & \textbf{8.03} & \textbf{13.68} & \textbf{8.77} \\ \hline
\end{tabular}
\label{result}
\end{minipage}}
\vspace{-1em}
\end{table*}

\section{Experiment}
\subsection{Experimental Setup}

To ensure reproducibility, we use seven publicly available Mandarin OpenSLR datasets (Table \ref{datasets}), each treated as private data for a curator. Mandarin presents additional challenges due to segmentation and semantic-level ambiguity \cite{wei-etal-2025-asr}. Datasets with predefined splits retain their original training, validation, and test sets; for the remaining datasets, we randomly split them into 60\% training, 20\% validation, and 20\% test sets. The Character Error Rate (CER) is reported as a fine-grained metric well-suited to Mandarin.

Five pairs of $n$-gram and NN LMs are trained on SLR33, SLR38, SLR47, SLR62, and SLR93 to form the source models, with the average CER reported on their respective test sets. The remaining datasets, SLR18 and SLR68, are reserved for testing to evaluate the generalization of the trained models. The ASR system is built using the Kaldi toolkit \cite{povey2011kaldi} with its ``Chain'' model\footnote{https://kaldi-asr.org/doc/chain.html}. The acoustic model is a Time Delay Neural Network (TDNN) \cite{waibel1989phoneme}, the NN LM is an RNN \cite{mikolov2010recurrent}, and the $n$-gram LM is a tri-gram trained with SRILM \cite{stolcke2002srilm}. Experiments are conducted on hardware with an Intel Xeon 72-core CPU, an NVIDIA Tesla K80 GPU, and 314GB of memory. All source models share identical training criteria, and all merging methods use the same set of source models.

\begin{figure}
	\centering
    \includegraphics[width=\linewidth]{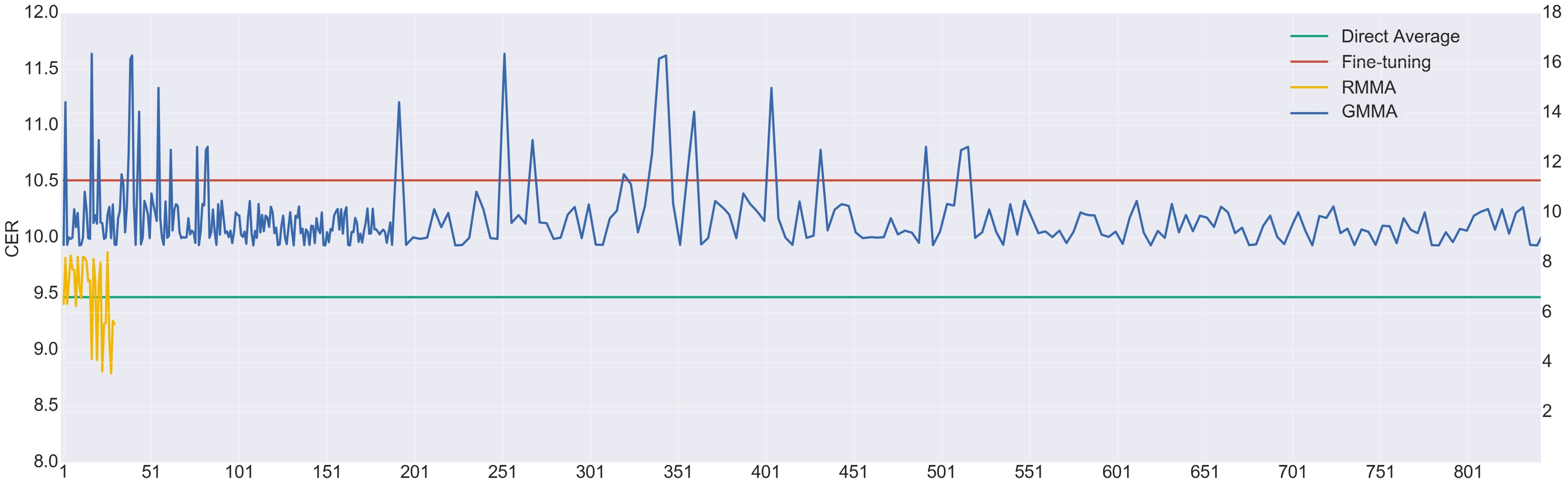}
    \vspace{-1.8em}
    \caption{Convergence behavior on training data. The secondary axis (right) corresponds to GMMA due to its larger variance.}
    \vspace{-1em}
	\label{efficiency}
\end{figure}

\subsection{Main Result}

Given the limited prior research on heterogeneous LM optimization, we present the first comprehensive comparison among: (a) Model fine-tuning, (b) Direct average of model parameters, (c) GMMA, (d) RMMA, and (e) Centralized model (trained on all five datasets, serving as a reference). Table \ref{result} reports CER evaluations on the test sets. The proposed RMMA achieves the lowest average CER, comparable to the centrally trained reference model, demonstrating the effectiveness of model merging. Evaluations on the SLR18 and SLR68 test sets further confirm RMMA’s superior generalization, followed by GMMA, both outperforming Direct Average and Fine-tuning. These results highlight the effectiveness of the proposed match-and-merge paradigm for optimizing heterogeneous LMs.

\subsection{Convergence Efficiency}
Figure \ref{efficiency} compares the convergence behaviors of the merging methods. When merging five source models, GMMA requires over 800 iterations and approximately 15 days to converge, whereas RMMA achieves convergence in fewer than 30 iterations within 2 days. RMMA shows a substantial CER reduction after just one iteration, while GMMA underperforms Direct Average and Fine-tuning during the first 60 iterations. These results highlight the practical advantage of RMMA's CER-based reward system, which guides search more efficiently than random exploration, achieving high model quality while being more efficient for large-scale model optimization.

\begin{table}[!t]
\centering
\vspace{-1em}
\caption{Performance evaluation (CER) of target models optimized on different numbers of source models.}
\vspace{1em}

\resizebox{0.91\columnwidth}{!}{
\begin{tabular}{|l|c|c|c|c|}
\hline
\textbf{Experiment} & \textbf{2 Models} & \textbf{3 Models} & \textbf{4 Models} & \textbf{5 Models} \\
\hline
\multicolumn{5}{|c|}{\textbf{Average}} \\ \hline
Direct Avg & 13.58 & 12.64 & 9.47 & 8.53 \\
\hline
RMMA & 9.56 & 9.23 & 8.91 & 8.03 \\ \hline
\multicolumn{5}{|c|}{\textbf{SLR18}} \\ \hline
Direct Avg & 21.54 & 19.57 & 15.23 & 14.98 \\
\hline
RMMA & 15.59 & 14.63 & 13.97 & 13.68 \\ \hline
\multicolumn{5}{|c|}{\textbf{SLR68}} \\ \hline
Direct Avg & 15.32 & 15.16 & 12.71 & 10.16 \\
\hline
RMMA & 14.65 & 14.27 & 10.39 & 8.77 \\
\hline
\end{tabular}
}
\vspace{-1em}
\label{tab:number}
\end{table}

\subsection{Performance Scaling with Number of Source Models}
We evaluate target model quality by varying the number of source models, added sequentially for performance comparison. Since the genetic algorithm and fine-tuning cannot process models sequentially, RMMA is compared only with the Direct Average baseline. As shown in Table \ref{tab:number}, increasing the number of source models generally improves performance, but RMMA achieves comparable quality using fewer models than Direct Average. The reinforcement learning agent in RMMA assigns higher weights to better models, mitigating the influence of lower-quality ones.

\section{Conclusion}
In this work, we introduce the optimization of heterogeneous language models for hybrid automatic speech recognition systems in a federated learning setting. To address the challenge of structural differences across multiple $n$-gram and neural network LM pairs, we formalize the Match-and-Merge paradigm and propose two novel algorithms: GMMA and RMMA. Experimental results show that both algorithms significantly improve LM performance, with RMMA achieving superior efficiency and performance nearly comparable to centralized training. These findings highlight the potential of federated learning combined with effective and efficient optimization strategies to preserve data privacy while maintaining high model quality, encouraging further exploration in this direction.

\subsection*{Acknowledgements}
The work was partially supported by the Research Grants Council (Hong Kong) (RGC) (PolyU25600624), the RCDTT Funding Body PolyU (UGC) (P0058790, 4-BBGV), the NSFC/RGC Joint Research Scheme (N\_PolyU5179/25), and the PolyU Start-up Fund (P0059983).

\bibliographystyle{IEEEbib}
\bibliography{refs}

\end{document}